\documentclass{article}

\usepackage{PRIMEarxiv}

\usepackage[utf8]{inputenc} 
\usepackage[T1]{fontenc}    
\usepackage{hyperref}       
\usepackage{url}            
\usepackage{booktabs}       
\usepackage{amsfonts}       
\usepackage{nicefrac}       
\usepackage{microtype}      
\usepackage{lipsum}
\usepackage{fancyhdr}       
\usepackage{graphicx}       
\graphicspath{{media/}}     

\usepackage{epsfig}
\usepackage{graphicx}
\usepackage{amsmath}
\usepackage{amssymb}
\usepackage{soul}
\usepackage{colortbl}
\usepackage{comment}
\usepackage{multirow}
\usepackage{diagbox}
\usepackage{slashbox}
\usepackage{float}
\usepackage[dvipsnames]{xcolor}

\newcommand{\bx}{\mathbf{x}}

\newcommand{\by}{\mathbf{y}}
\newcommand{\bc}{\mathbf{c}}

\title{Weakly-Supervised Dense Action Anticipation}

\author{
  Haotong Zhang \\
  Department of Mathematics\\
  National University of Singapore\\
  Singapore \\
  \texttt{haotongz@u.nus.edu} \\
   \And
  Fuhai Chen \\
  Department of Computer Science\\
  National University of Singapore\\
  Singapore \\
  \texttt{cfh3c@nus.edu.sg} \\
   \AND
   Angela Yao \\
   Department of Computer Science\\
   National University of Singapore\\
   Singapore \\
   \texttt{ayao@comp.nus.edu.sg} \\
}

\begin{document}
\maketitle

\begin{abstract}
Dense anticipation aims to forecast future actions and their durations for long horizons. 
Existing approaches rely on fully-labelled data, \emph{i.e.} sequences labelled with \emph{all} future actions and their durations. 
We present a (semi-) weakly supervised method using only a small number of fully-labelled sequences and predominantly sequences in which only the (one) upcoming action is labelled. 
To this end, we propose a framework that generates pseudo-labels for future actions and their durations and adaptively refines them through a refinement module. 
Given only the upcoming action label as input, these pseudo-labels guide action/duration prediction for the future. 
We further design an attention mechanism to predict context-aware durations. 
Experiments on the Breakfast and 50Salads benchmarks verify our method's effectiveness; we are competitive even when compared to fully supervised state-of-the-art models.
We will make our code available at: \url{https://github.com/zhanghaotong1/WSLVideoDenseAnticipation}. 
\end{abstract}

\section{Introduction}
Anticipating human actions is critical for real-world applications in autonomous driving, video surveillance, human-computer interaction, \emph{etc}.
According to the prediction horizons, the anticipation task is mainly investigated in two tracks: next-action anticipation~\cite{Vondrick16,Mahmud17, Qi17, Damen18, Farha18, Ke19, Fadime20} and dense anticipation~\cite{Farha18, Ke19, Fadime20}.
Next action anticipation predicts upcoming actions $\tau$ seconds in advance, where the value of $\tau$ is considered as 1 in many recent works. 
Dense anticipation predicts multiple actions into the future and their durations for long horizons of up to several minutes or an entire video.

\begin{figure}
\centering
\includegraphics[width=0.9\linewidth]{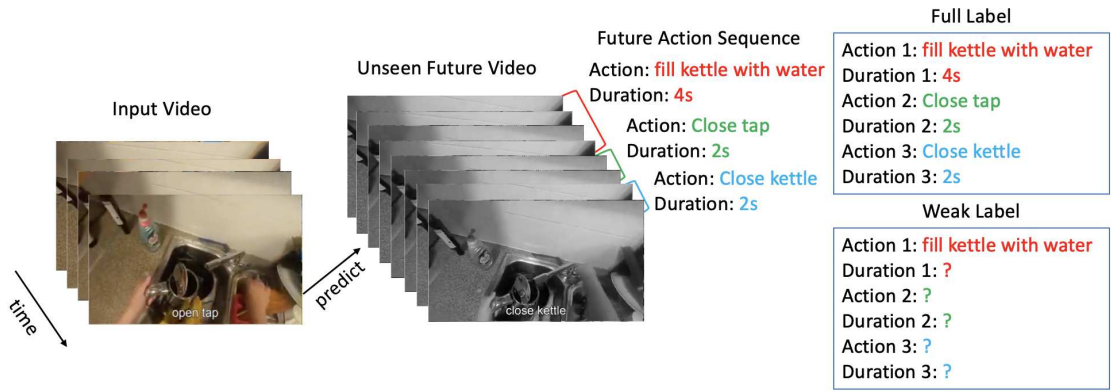}
\caption{
Dense anticipation with full supervision vs. weak supervision. The fully supervised label contains all the actions in the future video sequence as well as their durations.  In this work, we consider a weak label in which only the first action label without any duration information is available.  Our proposed framework is both semi- and weakly-supervised. We use a small set of fully-labelled videos, while the remainings are weakly-labelled.}
\label{fig:01_WS-VDA}
\end{figure}

Our paper focuses on the more challenging dense anticipation task where all existing methods~\cite{Farha18, Fadime20, Ke19} are fully supervised. 
Annotating videos for the fully supervised version of this task can be tedious, as it requires labelling the full set of actions in the subsequent sequence as well as their start and end times. In real-world videos, sequences are more likely to be labelled or tagged only at specific events. These tags are incomplete and instantaneous, \emph{i.e.} not present at every action and without duration information.
This motivates us to develop a weakly supervised dense anticipation framework that learns from video sequences with an incomplete set of action and duration labels. 
Specifically, we aim to learn from a small set of fully-labelled data and predominantly from 
weak labels in which the video segment is annotated only with the first action class of the anticipated sequence (see Fig.~\ref{fig:01_WS-VDA}).
This can greatly reduce the labelling effort as now we only need to provide the class label of a single action instead of all frames in the sequence.

In practice, this type of weak label is akin to the \emph{time-stamp annotations} used in weakly-supervised temporal action segmentation, in which an arbitrary frame from each action segment is labelled~\cite{Li21,Moltisanti19,Ma20}.
When annotating timestamps, annotators quickly go through a video and press a button when an action is occurring. This is $\sim$6x faster than marking the exact start and end frames of action segments~\cite{Ma20} and still provides strong cues to learn effective models for action segmentation.

In our case, our weak label can be viewed as an incomplete version of the full label since it has only one (the first) of the full set of action labels, and no duration labels.  Since each action label and action duration are all treated as separate terms in the loss for conventional anticipation methods~\cite{Mahmud17,Farha18,Fadime20}, a naive route to learn would be to ignore any missing labels from the loss.
This option, while simple, does not fully leverage the data of the weakly-labelled set.  We opt instead to learn an auxiliary model to generate pseudo-labels for the missing labels. The use of pseudo-labelling has become popular in unsupervised and semi-supervised learning~\cite{Helmstetter18, Yang21, Meng19, Yu20} and has been successful for tasks like image classification~\cite{Ge19, Wu17, Wangc21, Fang20} and segmentation~\cite{Dong19, Yao21, Chang20}. Inspired by these works, we propose a framework for learning a primary and conditional module for (semi-) weakly-supervised dense action anticipation.  
The conditional module is learned on a small fully-labelled training set to generate pseudo-labels for a larger weakly-labelled training set.  The pseudo-labelled weak data is then applied to learn the primary anticipation module which will be used during inference.  

Directly learning on the outputs of an auxiliary model is often not better than learning on the limited set of provided labels as it does not add new knowledge into the system. The phenomenon is referred to as confirmation bias~\cite{Arazo20}; extending previous solutions such as label smoothing~\cite{Zhang17} or label sampling and augmentation~\cite{Zhu18, Berthelot19, Iscen19} is non-trivial for sequence data. As such, we introduce an adaptive refinement method which learns refined sequence labels based on the predictions of the primary and conditional module. 
In our experimentation, we have observed that the accuracy of dense anticipation is highly sensitive to having the correct duration prediction, especially in the earlier anticipated actions\footnote{Consider a ground truth sequence of AABBCCDD where each letter is the action of a frame; a prediction of AAAABBCCDD would score a mean-over-classes of only 0.25 since all B, C and D frames are misaligned.}.  We are therefore motivated to ensure that the anticipated durations are correct.  To that end, we introduce an additional duration attention module applicable to recursive dense anticipation methods~\cite{Farha18,Fadime20}.  We compute an attention score between the observed video context and the hidden representation at each prediction step to explicitly emphasize the correlations, which greatly improves the duration accuracy.

The contributions of this paper are summarized as follows: 

1. We explore a novel and practical weakly-supervised dense anticipation task and propose an adaptive refinement method to make the most of weakly-labelled videos while using only a small number of fully-labelled videos.

2. We propose an attention scheme for predicting the duration of the anticipated actions which better accounts for the action correlations.

3. Our semi-supervised framework is flexible and applicable to a variety of dense anticipation backbones. The duration attention scheme serves as a plug-and-play module to improve the performance of recursive anticipation methods. Evaluation on standard benchmarks shows that our weakly supervised learning scheme can compete with state-of-the-art fully supervised approaches.

\section{Related Work}
Action recognition is the hallmark task of video understanding. 
In standard action recognition settings, short, trimmed video clips are classified with action labels.  
In contrast, action anticipation is applied to longer, untrimmed video sequences and aims to predict future actions \emph{before} they occur.
The task in next action anticipation is to predict the upcoming action  $\tau$ before it occurs.  
Various architectures ranging from recurrent neural networks (RNNs) ~\cite{Pirri19,Furnari20,Zhang20,Canuto20},  convolutional networks combined with RNNs~\cite{Mahmud17},  to transformers~\cite{Wang21} are proposed. 
The main focus of these works is to extract relevant information from the observations to predict the label of the action starting in $\tau$ seconds, varying between zero ~\cite{lan2014hierarchical} to 10s of seconds \cite{koppula2015anticipating}. 
Other models leverage external cues such as hand movements to help with the anticipation task~\cite{Liu20eccv, Dessalene21}.

Dense action anticipation predicts \emph{all} subsequent actions and their durations for longer horizons of the unobserved sequence. 
Recursive methods~\cite{Farha18, Fadime20} use an encoder to extract visual features from the observed sequence and use an RNN as a decoder to predict future actions and their duration sequentially.  As recursive predictions may accumulate and propagate errors, Ke~\emph{et al.}~\cite{Ke19} anticipates actions directly for specific future times in a single shot.  
When it comes to duration anticipation, all previous methods are relatively simple in that they apply a linear layer on top of the features of observed or predicted actions. Only past action features are used, without taking action correlations into account. Intuitively, actions with higher correlations with current action tend to influence more on current action's duration. Consequently, our method improves on previous works by introducing an attention mechanism for duration anticipation.

To date, all methods for dense anticipation~\cite{Farha18,Fadime20,Ke19} follow a fully supervised setting and require extensive annotations for learning.  
Driven by the laborious demand of fully labelled data in computer vision, some researchers focus on weakly- or semi-supervised learning to reduce annotation workload~\cite{Ahn19, Liu20, Lee21, Chen20}.  
Previously,~\cite{Ng20} apply a weakly-supervised model on forecasting future action sequences, where only action sequences rather than frame-wise labels are provided as coarse labels. They combine the attention scheme with GRU to recurrently predict action labels with more focus on related observed actions, which is similar to our duration attention.
Our work is similar in spirit to the teacher-student model~\cite{Tarvainen17, Laine16} which also uses an auxiliary model to support training.  However, we do not explicitly enforce label consistency between the two models and instead use a third refinement module to directly improve the pseudo-labels. 
Pseudo-labels are widely used in weak supervision~\cite{ChangY20, WangJ20, Zhang21}. Most often they are only propagated for unlabelled or semi-labelled data. We also generate for fully-labelled data and by minimizing the distance between ground truth and improved pseudo labels by our refinement module, we make the model adaptive refine the accuracy of pseudo labels.

\begin{figure*}[t!]
\centering
\includegraphics[width=1.0\linewidth]{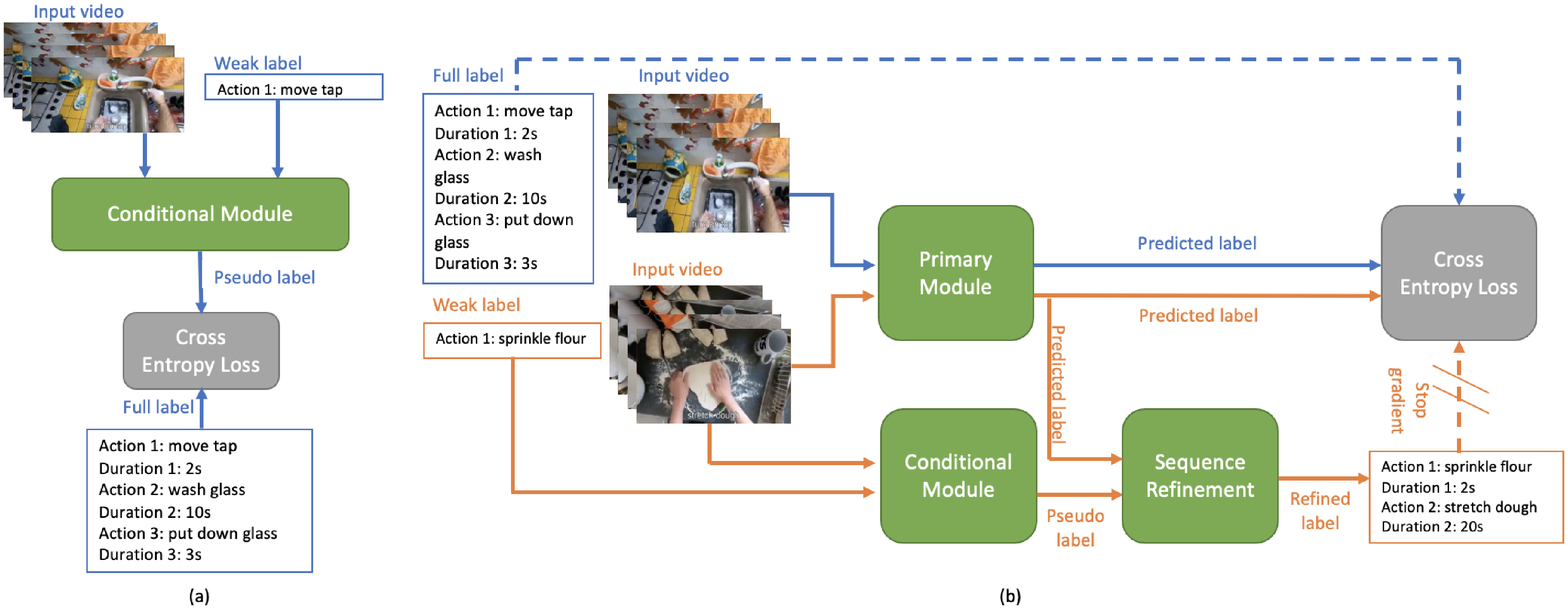}
\caption{Method overview. (a) The conditional module is trained on the small set of fully-labelled data to generate pseudo-labels. Once trained, it remains fixed, and does not contribute gradients in the following steps. (b) The primary module is trained on the small set of fully-labelled data and the large set of weakly-labelled data with the first future action label as the incomplete label. The weak labels are augmented into full pseudo-labels by refining the outputs of the conditional module.} 
\label{fig:02_overview}
\end{figure*}

\section{Method}

Our proposed framework is both semi- and weakly-supervised.  It is trained on a small set of fully labelled videos and a large set of weakly labelled videos with only the first action in the anticipated sequence. The model is comprised of three components: a primary module used during inference (Sec.~\ref{sec:primary}), a conditional module for generating pseudo-labels (Sec.~\ref{sec:conditional}), and a sequence refinement module (Sec.~\ref{sec:refinement}) to refine the estimated pseudo-labels.

We treat the primary and conditional modules as black-box encoder-decoders, where the observed video is encoded into features, while the decoder generates the anticipated action labels and durations.  
As the proposed framework is general, we can use any previously proposed dense anticipation model~\cite{Farha18,Fadime20,Ke19} as a backbone.  The training procedure can be broken down into two stages. The conditional module is trained initially on the fully-labelled set $\mathcal{F}$ so that it can be used to generate pseudo-labels for the weakly-labelled data.  The combined set of the fully-labelled and the pseudo-labelled weak data $\mathcal{W}$ then merged to train the primary module.  
Directly using the pseudo-labels may result in confirmation bias, as these labels are generated from a model which is learned only on the small set of fully-labelled data.  Therefore we refine the pseudo-labels with a sequence refinement module which is learned simultaneously with the primary module. Fig.~\ref{fig:02_overview} illustrates an overview.

\subsection{Preliminaries}\label{sec:prelim}
For a given video, $\textbf{x}=\{x_{1}, \dots, x_{t}, \dots, x_{T}\}$ denotes the set of $T$ observed frames. Dense anticipation aims to predict the future $M$ action labels $\textbf{c}=\{c_{1}, \dots, $ $c_{m}, \dots, c_{M}\}$ and associated durations $\textbf{d}=\{d_{1}, \dots,  d_{m}, \dots, d_{M}\}$ for frames $T+1$ onwards until the end of the video sequence. Note that $t$ is a per-frame index in the video, while $m$ is a per-action index. A fully supervised setting is then associated with a set of data $\mathcal{F}=\left\{\left(\textbf{x}, \textbf{c}, \textbf{d}\right)\right\}$. We also denote the action labels and duration jointly by $\by = \{y_1, \dots y_m \dots y_M\}$, where $y_m = (c_m, d_m)$, and distinguish the ground truth and the corresponding predictions as ${y}_m$ and $\hat{y}_m$ respectively.

Under a weakly-supervised setting, we assume we are given the set
$\mathcal{W}=\left\{\left(\textbf{x},\textbf{c}'\right)\right\}$, \emph{i.e.} observed videos of $T$ observed frames $\textbf{x}=\{x_{1}, \dots, x_{t}, \dots, x_{T}\}$, along with the weak label $\textbf{c}'=c_1$, \emph{i.e.} the action label of frame $x_{T+1}$.  There are no assumptions on $T$, \emph{i.e.} if the observed sequence end in the middle of an action, $c_1$ will be the current action label; if $T$ is exactly the last frame of an action, then $c_1$ will be the label of the next action.
This translates to the dense anticipation protocol of previous works~\cite{Fadime20,Farha18,Ke19} in which the first $X\%$ of a video is observed and predictions are made on the following $Y\%$ from $X$ to $X+Y$.  Therefore, we use as the weak label the first frame of the remaining $Y\%$.

We formulate dense anticipation as a mixed classification and regression task to anticipate the action labels and duration respectively.  Without any assumption on the backbone anticipation method, we will refer to the primary module as function $f(\bx)$ and the conditional module as function $f_{\text{cond}}(\bx, \bc')$.  The conditional module is trained based on the loss in Eq.~\ref{eq:condloss}. Then, $\mathcal{F}$ and $\mathcal{W}$ are used to train the primary module with conditional module fixed as elaborated in Section ~\ref{sec:primary}. The main issue is how to adjust pseudo-labels.

\subsection{Conditional Module}~\label{sec:conditional}
The conditional module $\tilde{\by} = f_{\text{cond}}(\textbf{x},\textbf{c}')$ is an auxiliary component trained for generating pseudo labels $\tilde{\by}$ for the weak set $\mathcal{W}$.  To do so, it is trained in the standard way using $\mathcal{F}$ with the following loss function
\begin{equation}
    L_{\text{cond}} = \frac{1}{|\mathcal{F}|}\sum_{\mathcal{F}}\sum_{m=1}^M\left( -c_m\log(\hat{c}^{\text{cond}}_m) + (d_m-\hat{d}^{\text{cond}}_m)^2\right),
\label{eq:condloss}
\end{equation}
where the first term is a cross-entropy loss for the anticipated action label $\hat{c}^{\text{cond}}_m$, while the second term is an MSE for the predicted action duration $\hat{d}^{\text{cond}}_m$.  After training, the conditional module remains fixed.  For generating pseudo labels, we simply apply $f_{\text{cond}}$.  However, to make full use of the weak label, we replace the estimated $\hat{c}_1$ with the weak label  $\bc' = c_1$, \emph{i.e.}
$\tilde{\by}= \{(c_1, \hat{d}^{\text{cond}}_1), (\hat{c}^{\text{cond}}_2,\hat{d}^{\text{cond}}_2), \dots,  (\hat{c}^{\text{cond}}_M,\hat{d}^{\text{cond}}_M)\}$.

\subsection{Primary model}~\label{sec:primary}
The primary module $\hat{\by} = f(\textbf{x})$ predicts the future action and duration sequence $\hat{\by}$ given video $\bx$ and is the module used for inference.  During training, the objective is to minimize a loss based on the labelled ground truth $\mathcal{F}$ and the refined pseudo-labels of $\mathcal{W}$, \emph{i.e.}
\begin{equation}
\begin{split}
    L_{\text{prim}} = & \frac{1}{|\mathcal{F}|}\sum_{\mathcal{F}}\sum_{m=1}^M \left(-c_m\log(\hat{c}_m) + (d_m-\hat{d}_m)^2\right) + \frac{1}{|\mathcal{W}|}\sum_{\mathcal{W}}\left(-c_1\log(\hat{c}_1)\right)\\
    +&\frac{1}{|\mathcal{W}|}\sum_{\mathcal{W}}\left(\sum_{m=2}^M \left(-\tilde{c}_m'\log(\hat{c}_m)\right) + \sum_{m=1}^M(\tilde{d}_m'-\hat{d}_m)^2\right),
\label{eq:primloss}
\end{split}
\end{equation}
where $\hat{y}_m = (\hat{c}_m, \hat{d}_m)$ is the predicted label from the primary module while $\tilde{y}_m' = (\tilde{c}_m', \tilde{d}_m')$ is the refined pseudo-labels (see Sec.~\ref{sec:refinement}) on $\mathcal{W}$.  The first two terms represent the loss based on ground truth labels on $\mathcal{F}$ and $\mathcal{W}$; we term this $L_{\text{label}}$.  
The third term in the loss is based on pseudo-labels on $\mathcal{W}$ and we term this $L_{\text{pseudo-label}}$.

\subsection{Sequence Refinement}\label{sec:refinement}

Directly using the pseudo-labels from the conditional module to train the primary module does not allow us to fully benefit from $\mathcal{W}$, since the conditional module is trained only on $\mathcal{F}$. As $\mathcal{F}$ is quite small (5-15\% of the training set in our case), there is also the risk of confirmation bias~\cite{Arazo20}. 
To mitigate this possibility, we learn a refinement module to refine the pseudo-labels from the conditional module. 
For a video $\bx$, the refinement module can be expressed as a function $F$ applied the predicted labels from the primary module and the estimated pseudo-labels from the conditional module, \emph{i.e.}
\begin{equation}
    \tilde{\by}' = F(\hat{\by}, \tilde{\by}) = F \left(f(\textbf{x}), f_{\text{cond}}((\textbf{x},\textbf{c}'))\right).
\end{equation}
\noindent We propose two refinement schemes as different options for $F$ which we outline below.\\

\noindent \textbf{Linear Refinement.} 
As a naive baseline, we first propose to use a weighted geometric mean of the primary and conditional module outputs, where we consider $\hat{c}$ as a probability estimate over the classes.  To that end, the refined label can be defined as 

\begin{equation}
    \tilde{\by}' = f(\textbf{x})^{\frac{1}{\alpha+1}} \cdot f_{\text{cond}}((\textbf{x}, \textbf{c}'))^{\frac{\alpha}{\alpha+1}},
\label{eq:klmin}
\end{equation}
where $\alpha$ is a hyperparameter determining the weighting of each component. 
Note that $\tilde{\by}'$ is actually the optimal solution when considering a linear weighting of the minimal KL divergences between (1) the refined pseudo-label $\tilde{\by}'$ and the estimate of the primary module $\hat{\by}$ as well as between $\tilde{\by}'$ and (2) the estimate of the conditional module $\tilde{\by}$.  
Intuitively, the refined output is the ``closest'' sequence to both modules' predictions.

From Eq.~\ref{eq:klmin}, it can be observed that when $\alpha=\infty$, $\tilde{\by}'=f_{\text{cond}}(\textbf{x}, \textbf{c}')$ while $\alpha=0$ gives $\tilde{\by}'=f(\textbf{x})$. These two extreme cases correspond to the refinement directly using the conditional or primary module outputs as the refined sequence respectively.  We define a schedule for $\alpha$ to decrease from a large to a small value.  This is based on the rationale that at the outset of training, the primary module is not so accurate and will need to rely on the conditional module, but as training progresses a smaller $\alpha$ is more suitable.\\

\noindent \textbf{Adaptive Refinement.}
Instead of a manually set schedule for $\alpha$, we can also directly learn a refined output.  Ideally, we would like for the refined outputs $\tilde{\by}'$ to be more accurate than the outputs of both $f(\textbf{x})$ and $f_{\text{cond}}(\textbf{x}, \textbf{c}')$.  We can do this by leveraging the ground truth labels of $\mathcal{F}$ and adding a loss on the refined output $\tilde{\by}'$:
\begin{equation}
\begin{split}
    L_{\text{adap}} = L_{\text{prim}} + 
    \frac{1}{|\mathcal{F}|}\sum_{\mathcal{F}}\sum_{m=1}^M\bigg(-c_m\log(\tilde{c}'_m) + (d_m-\tilde{d}'_m)^2\bigg).
\end{split}
\label{eq:autoloss}
\end{equation}

\noindent The adaptive refinement is realized via a linear layer that takes predicted and pseudo sequences and outputs a refined one.
One key change made when learning the adaptive refinement as opposed to the linear refinement is that the conditional module is trained on only a portion of $\mathcal{F}$ (we opt for half out of simplicity). We purposely limit the training of the conditional module to prevent the refinement module from fully relying on its output.  
Then, $\mathcal{F}$ is used to train the primary and refinement module simultaneously. The objective function contains two parts: loss between output from the primary module $\hat{y}$ and ground truth (\emph{i.e.} the first term in Eq.~\ref{eq:primloss}) and refined output $\tilde{y}'$ and ground truth (\emph{i.e.} the second term in Eq.~\ref{eq:autoloss}).
Lastly, $\mathcal{F}$ as well as $\mathcal{W}$ is then applied to learn the primary and refinement module concurrently based on the loss in Eq.~\ref{eq:autoloss}.  
We refer readers to Supplementary Section 8 to get a better idea of the training process.

\subsection{Duration Attention}\label{sec:attn}

We introduce attention for the duration estimation; this is applicable only to recursive dense anticipation methods~\cite{Farha18,Fadime20}.  At the decoder, the action label and duration for action $m$ would be classified and regressed directly from the hidden state $H_m$. We propose to add an attention score between the hidden state and the input video to improve the duration estimate.   Specifically, given video encoding $\mathcal{I}$, the attention weighted sum of the encoding can be defined as:

\begin{equation}\label{eq:attn}
    \mathbf{attn}(H_{m}',\mathcal{I}) = \text{softmax}(\frac{H_{m}'\mathcal{I}^{\intercal}}{\sqrt{d_I}})\mathcal{I}, \qquad \text{where} \qquad H_{m}' = WH_{m}+b 
\end{equation}
\noindent where $W \in \mathbb{R}^{d_I\times d_h}$ and $b \in \mathbb{R}^{d_I}$ are learned parameters, $\mathcal{I}^{\intercal}$ is the transpose of $\mathcal{I}$, $d_h$ and $d_I$ are the dimensionality of $H_m$ and $\mathcal{I}$ respectively. The attention-based duration $\hat{d}_m$ is estimated as a linear transformation of the previous hidden state $H_{m-1}$ and the weighted encoding:
\begin{equation}\label{eq:duration}
    \hat{d}_{m} = [\mathbf{attn}(H_{m}',\mathcal{I}), H_{m-1}]\mathbf{\beta}+\mathbf{\epsilon}
\end{equation}

\noindent where $\mathbf{\beta}$, $\mathbf{\epsilon}$ are learned parameters and $[ \cdot ]$ denotes a concatenation.\\

\noindent \textbf{Duration Attention Regularizer.} To further minimize the prediction differences between the primary and conditional module, we encourage the attention score between the two modules to be similar.  To that end, we add to the objective functions Eq.~\ref{eq:primloss} and Eq.~\ref{eq:autoloss} an $l_2$-norm between the attention scores of the conditional and primary modules, \emph{i.e.} 
\begin{equation}~\label{eq:duration_regularizer}
\begin{split}
    L_{\text{prim}}' = 
    L_{\text{prim}}
    + \sum_{m=1}^M\|\mathbf{attn}_m^{\text{prim}}-\mathbf{attn}_m^{\text{cond}}\|_2^2 
\end{split}
\end{equation}
where $\mathbf{attn}_m^{\text{prim}}$ and $\mathbf{attn}_m^{\text{cond}}$ represent the attention scores of step $m$ in the primary and conditional modules respectively.  The same regularizer is also added to 
Eq.~\ref{eq:autoloss} to yield $L_{\text{adap}}'$.

\section{Experiments}

\subsection{Datasets, Evaluation \& Implementation Details}

We evaluate our method on the two benchmark datasets used in dense anticipation: Breakfast Actions~\cite{Kuehne14} and 50Salads~\cite{Stein13}. Both datasets record realistic cooking activities, with each video featuring a sequence of continuous actions in making either a breakfast item or a salad\footnote{Dataset details are in the Supplementary Section 1.}.  From the designated training splits of each dataset, we partition 15\% and 20\% of the training data for the fully labelled set $\mathcal{F}$  for Breakfast and 50Salads respectively\footnote{We use a slightly higher percentage for 50Salads due to the small dataset size}.  The remaining 85\% / 80\% of training sequences are assigned to $\mathcal{W}$ and have only a weak label, \emph{i.e.} the single action label $c_1$ (see Sec.~\ref{sec:prelim}).  Following the conventions of~\cite{Farha18,Fadime20,Ke19}, we observe 20\% or 30\% of the video and anticipate the subsequent 20\% and 50\% of the video sequence (with additional results on 10\% and 30\% in the Supplementary Section 2).  In line with previous works, we evaluate our anticipation results with mean over classes (MoC)~\cite{Farha18}.  

As input features, we use the 64-dimension Fisher vectors computed on top of improved dense trajectories~\cite{IDT} as provided by~\cite{Farha18} on a per-frame basis.  

Currently, as all dense anticipation methods are fully supervised, there are no direct comparisons to competing state-of-the-art methods.  However, as our framework is general, we experiment with 3 different anticipation methods as backbones in a series of self-comparisons.  We test using (1) a naive RNN where both encoder and decoder is a one-layer LSTM with 512 hidden dimensions (2) the one-shot method of Ke~\cite{Ke19} and (3) the recursive method of Sener~\cite{Fadime20}.  
Our result for Ke \emph{et al.} is our re-implementation as they do not provide source code; our fully-supervised re-implementation yields similar values as their reported results.
All hyperparameters follow the original settings in their papers.

For the linear refinement method, $\alpha$ begins from 30 and decreases to 0.5 with a decay rate of 0.95 per epoch. 
The batch size is 2 for 50Salads and 16 for Breakfast. 
Using linear refinement, the model converges at about 20 epochs for the first step and 25 epochs for the second step.  The model converges at about 15 epochs for the first step, 20 for the second and third step when using adaptive refinement.

\begin{table*}[t!]
\caption{MoC of different models. Results reported in Baseline (1) for Ke~\cite{Ke19} and Sener~\cite{Fadime20} are taken directly from their published results. Other results are averaged on the officially provided different splits for training (which is further split into fully- and weakly-labeled sets randomly according to the percentages mentioned above) and test set.}
\begin{center}
\small
\begin{tabular}{|p{1.3cm}|p{0.7cm}|p{0.7cm}|p{0.7cm}|p{0.7cm}|p{0.7cm}|p{0.7cm}|p{0.7cm}|p{0.7cm}|}
\hline
\quad & \multicolumn{4}{|c|}{Breakfast} &\multicolumn{4}{|c|}{50Salads} \\
\hline
Obs. & \multicolumn{2}{|c|}{20\%} & \multicolumn{2}{|c|}{30\%} & \multicolumn{2}{|c|}{20\%} & \multicolumn{2}{|c|}{30\%} \\
\hline
Pred. & 20\% & 50\% & 20\% & 50\% & 20\% & 50\% & 20\% & 50\% \\
\hline
\rowcolor{green!20}
\multicolumn{9}{|l|}{Baseline 1: $f(\bx)$, fully-supervised on entire training set (theoretical upper bound)}\\
\hline
\rowcolor{green!20}
RNN & 6.53 & 5.30 & 8.52 & 5.37 & 9.71 & 7.82 & 12.64 & 8.54\\
\rowcolor{green!20}
Ke~\cite{Ke19} & 11.92 & 7.03 & 12.26 & 8.18 & 11.53 & 9.50 & 15.92 & 9.89\\
\rowcolor{green!20}
Sener~\cite{Fadime20} & 13.10 & 11.10 & 17.00 & 15.10 & 19.90 & 15.10 & 22.50 & 11.20\\
\hline
\rowcolor{red!20}
\multicolumn{9}{|l|}{Baseline 2: $f(\bx)$, supervised on full label set $\mathcal{F}$ (theoretical lower bound)}  \\
\hline
\rowcolor{red!20}
RNN & 3.92 & 2.35 & 5.48 & 4.26 & 8.08 & 5.45 & 8.13 & 6.70\\
\rowcolor{red!20}
Ke~\cite{Ke19} & 6.81 & 5.39 & 7.32 & 5.88 & 8.36 & 4.51 & 11.19 & 8.23\\
\rowcolor{red!20}
Sener~\cite{Fadime20} & 6.19 & 4.90 & 7.30 & 5.92 & 8.67 & 7.01 & 12.73 & 8.00\\
\hline
\rowcolor{cyan!20}
\multicolumn{9}{|l|}{Baseline 3: $f(\bx)$, supervised on full label set $\mathcal{F}$ + weak set $\mathcal{W}$ with $L_{\text{label}}$} \\
\hline
\rowcolor{cyan!20}
RNN & 6.01 & 4.29 & 7.56 & 5.93 & 9.33 & 6.96 & 11.45 & 8.54\\
\rowcolor{cyan!20}
Ke~\cite{Ke19} & 8.89 & 5.71 & 10.05 & 7.59 & 9.25 & 6.11 & 13.17 & 9.80\\
\rowcolor{cyan!20}
Sener~\cite{Fadime20} & 7.64 & 5.54 & 8.05 & 6.77 & 9.97 & 7.89 & 13.30 & 9.61\\
\hline
\rowcolor{blue!20}
\multicolumn{9}{|l|}{Our model with adaptive refinement but without duration attention.} \\
\hline
\rowcolor{blue!20}
RNN & 7.85 & 7.96 & 8.33 & 8.21 & 10.48 & 7.40 & 13.04 & 10.05\\
\rowcolor{blue!20}
Ke~\cite{Ke19} & 9.74 & 6.24 & 11.02 & 9.24 & 11.84 & 9.27 & 13.88 & 12.81\\
\rowcolor{blue!20}
Sener~\cite{Fadime20} & 8.98 & 7.71 & 9.71 & 7.31 & 12.62 & 9.44 & 13.94 & 10.73\\
\hline
\multicolumn{9}{|l|}{Our full model with adaptive refinement and duration attention.} \\
\hline
RNN & 9.12 & 8.33 & 10.17 & 8.90 & 12.11 & 9.57 & 14.37 & 10.91\\
Sener~\cite{Fadime20} & 9.74 & 8.56 & 11.63 & 8.99 & 12.41 & 9.67 & 14.94 & 12.14\\
\hline
\end{tabular}
\end{center}
\label{tb:result}
\end{table*}

\subsection{Supervised Baselines}
We first compare the impact that the amount of data would have on the fully supervised case (see Table~\ref{tb:result}).  We design three baselines and in each case, train a stand-alone primary module.  Baseline (1) is fully supervised on the entire training set -- this signifies the upper bound that our weakly-supervised method can achieve.
Baseline (2) is supervised on only the labelled set $\mathcal{F}$.  This baseline gives some indicator of the accuracy of the conditional module before the weak label is applied to replace $\hat{c}_1$ and acts as a lower bound.  Baseline (3) supervised on the given labels of $\mathcal{F}$ and $\mathcal{W}$, \emph{i.e.} applying the first two terms or $L_{\text{label}}$ of Eq.~\ref{eq:primloss}.  This baseline tells us what can be learned from the full set of provided labels.  

Full supervision with the entire training set, \emph{i.e.} Baseline (1) achieves the best results, with the model of Sener~\emph{et al.}~\cite{Fadime20} performing best.  However, performance drops with fewer labels, \emph{i.e.} Baselines (2) and (3) and the one-shot method of Ke~\emph{et al.}~\cite{Ke19} is slightly stronger than~\cite{Fadime20}.  The gains from adding the labels of the weak set $\mathcal{W}$, \emph{i.e.} from Baseline (2) to (3) demonstrate that having even a single $c_1$ label helps to improve MoC by 1-2\%.

\subsection{Impact of Adding Pseudo-Labels and Duration Attention}
If we add pseudo-labels to train the primary module, \emph{i.e.} by applying the full loss given in Eq.~\ref{eq:primloss} (see Table~\ref{tb:result}, fourth section) and using adaptive refinement, we observe that we gain in performance across the board when compared to Baseline (3), even though it uses the same amount of provided ground truth labels. 
The most impressive is the RNN encoder-decoder model.  With only the pseudo-labels from the weak set, we can surpass the original fully supervised baseline.  Using the one-shot method from Ke~\cite{Ke19}, we can surpass the supervised baseline when anticipating 50\% of the sequence after observing 30\% for both Breakfast and 50Salads.  
On Sener's model~\cite{Fadime20}, however, we are not able to surpass the fully supervised baseline, though the gap closes progressively. Our full model (Table~\ref{tb:result}, fifth section) which incorporates the attention duration sees additional gains in most settings. 
There is also a visual explanation in Supplementary Section 4 which intuitively illustrates different correlations between different observed actions and current predicted action.
Note that we do not apply the duration attention to the model of Ke~\cite{Ke19} since it is not recursive. 

All three backbones improve from Baseline (3) when adding adaptive refinement and duration attention. Given the challenge of the dense anticipation task, however, the overall performance is still very low, especially for the simple RNN and Ke's~\cite{Ke19} model.  This is likely the reason why adding our framework can outperform the fully supervised case. As the models are rather simplistic, we speculate they 
cannot fully leverage all the ground truth labels from the entire training dataset (Table~\ref{tb:result}, Baseline (1)). Training with our framework (Table~\ref{tb:result}, our model in purple and white section) may result in even higher accuracies because our refined pseudo-labels, while less accurate than ground truth, model a simpler distribution.

\subsection{Future Horizon of Anticipated Actions}

We analyze in Table~\ref{tb:action_acc} the anticipated actions over time by computing the accuracy for the first future action (weak label) versus the next three actions (no label).
The trends for the two settings are very different; Baseline 3 without the conditional module has a sharp drop-off from the second action.  This is unsurprising since most videos have only a weak label of the first future action.  Incorporating our conditional module with the refined pseudo-labels improves the first action's accuracy and decreases the drop-off of subsequent actions. 
Refer to Supplementary Section 7 for a visualization of the anticipated action sequence.

\begin{table}[H]
\caption{Accuracy of the predicted actions at different time steps.}
\centering
\begin{tabular}{|c|c|c|c|c|}
\hline
\quad & First  & Second & Third & Fourth\\
\hline
Baseline 3
& 16.17 & 6.49 & 3.22 & 1.67\\
Our full model  & 18.75 & 14.33 & 9.09 & 5.49\\
\hline
\end{tabular}
\label{tb:action_acc}
\end{table}

\subsection{Ablation Study}

In the following experiments we use Sener's~\cite{Fadime20} method as the backbone, an observation of 30\% and anticipation of 10\%.  Table~\ref{tb:corr} verifies that refining the pseudo-labels is more effective than training with them directly.  
Furthermore, the learned adaptive refinement is better than the linear refinement as it improves upon the linear scheme by 4\% on both datasets. 

In addition to Fisher vector IDT features, we also experiment with the ground truth labels and the stronger I3D features~\cite{Carreira17} as inputs, the result is shown in Table~\ref{tb:feat}. 
To use ground truth labels as input, we simply use a one-hot vector.
It gives much higher accuracy, indicating that there is still some gap in recognition performance.
The same gap was also confirmed in~\cite{Fadime20}. In line with previous results which use both features, I3D achieves higher MoC than Fisher vector.  We observe, however, that using I3D features requires longer training time, 
\emph{i.e.} 20 epochs in step 1 and 2, 25 epochs in step 3 
(we refer readers to Section 3.4 in the main paper and Section 8 in the Supplementary for a detailed training procedure), 
likely due to the larger dimensionality of I3D compared to Fisher vectors.

\begin{table}[H]
\quad
\parbox{0.45\linewidth}{
\caption{MoC on different refinements.}
\centering
\begin{tabular}{|c|c|c|}
\hline
\quad & Breakfast & 50Salads \\
\hline
No refinement & 6.28 & 10.31\\
Linear & 7.79 & 12.17 \\
Adaptive & 12.78 & 16.24 \\
\hline
\end{tabular}
\label{tb:corr}
}
\quad
\parbox{0.45\linewidth}{
\caption{MoC on different video features.}
\centering
\begin{tabular}{|c|c|c|}
\hline
\quad & Breakfast & 50Salads \\
\hline
Ground truth & 61.30 & 35.40 \\
Fisher vector & 12.78 & 16.24 \\
I3D & 15.65 & 21.30 \\
\hline
\end{tabular}
\label{tb:feat}
}
\end{table}

\section{Conclusion}

In this paper, we investigate a novel dense anticipation task, emphasizing pseudo labels to promote anticipation accuracy using weakly-labelled videos. To predict accurate action/duration sequences, we propose a sequence refinement method that generates pseudo sequences conditioned on the next-step action and adaptively refines the pseudo sequences to guide prediction. 
We also introduce duration attention which takes action correlations into account to boost duration anticipation.
The proposed method outperforms, if not better than, other fully supervised methods while requiring far less annotation effort.
More datasets will be involved in future works.

\noindent\textbf{Acknowledgements }This research is supported by the National Research Foundation, Singapore under its NRF Fellowship for AI (NRF-NRFFAI1-2019-0001).

\bibliographystyle{unsrt} 
\bibliography{anticipation}

\end{document}


\maketitle

\section{Dataset Details}

\noindent \textbf{Breakfast Actions} contains 1712 videos which are performed by 52 different individuals in 18 different kitchens. 
The videos are unscripted and uncontrolled with natural lighting, view points and environments. 
\textbf{50Salads} is food preparation dataset capturing 25 people preparing 2 mixed salads each. 
Both datasets have standardized train-test splits which we follow.  We further split the training set into fully- and weakly-labelled sets, with specific proportions and other details in Table~\ref{tb:info}.

\begin{table}[H]
\caption{Basic information of dataset.}
\begin{center}
\small
\begin{tabular}{|c|c|c|c|c|c|c|c|c|}
\hline
Dataset & fps & \tabincell{c}{Video duration\\median, mean$\pm$std} & Classes & Total & Train & Full & Weak & Test \\
\hline
Breakfast & 15 & 91s, 140s$\pm$122 & 48 & 1712 & 1460& 15\% & 85\% & 252\\
50Salads & 30 & 389s, 370s$\pm$106 & 19 & 50 & 40 & 20\% & 80\% & 10\\
\hline
\end{tabular}
\end{center}
\label{tb:info}
\end{table}

\section{Complete Results}

\begin{table*}
\caption{MoC of different methods on Breakfast. Better viewed in colour.}
\begin{center}
\small
\begin{tabular}{|p{1.3cm}|p{0.7cm}|p{0.7cm}|p{0.7cm}|p{0.7cm}|p{0.7cm}|p{0.7cm}|p{0.7cm}|p{0.7cm}|}
\hline
Observed & \multicolumn{4}{|c|}{20\%} & \multicolumn{4}{|c|}{30\%} \\
\hline
Predicted & 10\% & 20\% & 30\% & 50\% & 10\% & 20\% & 30\% & 50\% \\
\hline
\rowcolor{green!20}
\multicolumn{9}{|l|}{Baseline 1: $f(\bx)$, fully-supervised on entire training set (theoretical upper bound)} \\
\hline
\rowcolor{green!20}
RNN & 8.39 & 6.53 & 5.93 & 5.30 & 9.19 & 8.52 & 7.92 & 5.37 \\
\rowcolor{green!20}
Ke & 13.04 & 11.92 & 7.76 & 7.03 & 14.24 & 12.26 & 11.60 & 8.18 \\
\rowcolor{green!20}
Sener & 15.60 & 13.10 & 12.10 & 11.10 & 19.50 & 17.00 & 15.60 & 15.10 \\
\hline
\rowcolor{red!20}
\multicolumn{9}{|l|}{fully-supervised on the fully labelled subset (theoretical lower bound)} \\
\hline
\rowcolor{red!20}
RNN & 5.48 & 3.92 & 3.45 & 2.35 & 5.98 & 5.48 & 5.23 & 4.26 \\
\rowcolor{red!20}
Ke & 7.18 & 6.81 & 5.32 & 5.39 & 9.83 & 7.32 & 6.33 & 5.88 \\
\rowcolor{red!20}
Sener & 7.47 & 6.19 & 5.18 & 4.90 & 7.93 & 7.30 & 5.47 & 5.92 \\
\hline
\rowcolor{cyan!20}
\multicolumn{9}{|l|}{Baseline 3: $f(\bx)$, supervised on full label set $\mathcal{F}$ + weak set $\mathcal{W}$ with $L_{\text{label}}$} \\
\hline
\rowcolor{cyan!20}
RNN & 7.29 & 6.01 & 5.16 & 4.29 & 8.34 & 7.56 & 6.62 & 5.93 \\
\rowcolor{cyan!20}
Ke & 9.76 & 8.89 & 6.51 & 5.71 & 11.71 & 10.05 & 8.52 & 7.59 \\
\rowcolor{cyan!20}
Sener & 8.09 & 7.64 & 6.37 & 5.54 & 9.38 & 8.05 & 7.45 & 6.77 \\
\hline
\rowcolor{blue!20}
\multicolumn{9}{|l|}{Our model with adaptive refinement but without duration attention.} \\
\hline
\rowcolor{blue!20}
RNN & 9.87 & 7.85 & 6.89 & 7.96 & 10.90 & 8.33 & 8.31 & 8.21 \\
\rowcolor{blue!20}
Ke & 11.82 & 9.74 & 7.32 & 6.24 & 13.75 & 11.02 & 10.06 & 9.24 \\
\rowcolor{blue!20}
Sener & 9.03 & 8.98 & 7.64 & 7.71 & 10.11 & 9.71 & 8.11 & 7.31 \\
\hline
\multicolumn{9}{|l|}{Our full model with adaptive refinement and duration attention} \\
\hline
RNN & 9.93 & 9.12 & 8.70 & 8.33 & 12.55 & 10.17 & 9.54 & 8.90 \\
ours & 10.09 & 9.74 & 7.99 & 8.56 & 12.78 & 11.63 & 10.73 & 8.99 \\
\hline
\end{tabular}
\end{center}
\label{tb:breakfast}
\end{table*}

\begin{table*}
\caption{MoC of different methods on 50Salads. Better viewed in colour.}
\begin{center}
\small
\begin{tabular}{|p{1.3cm}|p{0.7cm}|p{0.7cm}|p{0.7cm}|p{0.7cm}|p{0.7cm}|p{0.7cm}|p{0.7cm}|p{0.7cm}|}
\hline
Observed & \multicolumn{4}{|c|}{20\%} & \multicolumn{4}{|c|}{30\%} \\
\hline
Predicted & 10\% & 20\% & 30\% & 50\% & 10\% & 20\% & 30\% & 50\% \\
\hline
\rowcolor{green!20}
\multicolumn{9}{|l|}{Baseline 1: $f(\bx)$, fully-supervised on entire training set (theoretical upper bound)}\\
\hline
\rowcolor{green!20}
RNN & 11.49 & 9.71 & 9.60 & 7.82 & 12.97 & 12.64 & 11.83 & 8.54 \\
\rowcolor{green!20}
Ke & 12.29 & 11.53 & 10.97 & 9.50 & 16.34 & 15.92 & 11.56 & 9.89 \\
\rowcolor{green!20}
Sener & 25.50 & 19.90 & 18.20 & 15.10 & 30.60 & 22.50 & 19.10 & 11.20 \\
\hline
\rowcolor{red!20}
\multicolumn{9}{|l|}{Baseline 2: $f(\bx)$, supervised on full label set $\mathcal{F}$ (theoretical lower bound)} \\
\hline
\rowcolor{red!20}
RNN & 9.81 & 8.08 & 6.59 & 5.45 & 10.65 & 8.13 & 7.52 & 6.70 \\
\rowcolor{red!20}
Ke & 9.16 & 8.36 & 7.65 & 4.51 & 12.69 & 11.19 & 8.31 & 8.23 \\
\rowcolor{red!20}
Sener & 11.36 & 8.67 & 7.30 & 7.01 & 13.16 & 12.73 & 10.95 & 8.00 \\
\hline
\rowcolor{cyan!20}
\multicolumn{9}{|l|}{Baseline 3: $f(\bx)$, supervised on full label set $\mathcal{F}$ + weak set $\mathcal{W}$ with $L_{\text{label}}$} \\
\hline
\rowcolor{cyan!20}
RNN & 10.60 & 9.33 & 8.31 & 6.96 & 13.25 & 11.45& 10.55 & 8.54 \\
\rowcolor{cyan!20}
Ke & 11.87 & 9.25 & 8.83 & 6.11 & 14.97 & 13.17 & 10.74 & 9.80 \\
\rowcolor{cyan!20}
Sener & 12.91 & 9.97 & 8.86 & 7.89 & 14.63 & 13.30 & 11.19 & 9.61 \\
\hline
\rowcolor{blue!20}
\multicolumn{9}{|l|}{Our model with adaptive refinement but without duration attention.} \\
\hline
\rowcolor{blue!20}
RNN & 12.72 & 10.48 & 9.84 & 7.40 & 14.52 & 13.04 & 12.72& 10.05 \\
\rowcolor{blue!20}
Ke & 15.00 & 11.84 & 10.96 & 9.27 & 15.66 & 13.88 & 12.89 & 12.81 \\
\rowcolor{blue!20}
Sener & 13.07 & 12.62 & 10.01 & 9.44 & 15.25 & 13.94 & 11.44 & 10.73 \\
\hline
\multicolumn{9}{|l|}{Our full model with adaptive refinement and duration attention} \\
\hline
RNN & 14.53 & 12.11 & 10.06 & 9.57 & 15.09 & 14.37 & 13.25 & 10.91 \\
ours & 16.80 & 12.41 & 10.12 & 9.67 & 16.24 & 14.94 & 13.53 & 12.14 \\
\hline
\end{tabular}
\end{center}
\label{tb:salad}
\end{table*}

We provide a complete set of anticipations (10\%, 20\%, 30\% and 50\%) in Tables~\ref{tb:breakfast} and~\ref{tb:salad} for Breakfast and 50Salads respectively.  
Findings are consistent with the 20\% and 50\% results in the main paper. Baseline 1 is a fully supervised version; the MoC of Baseline 2 drops because we omit a large proportion of videos from the training set. We observe an increase in the performance of Baseline 3 compared to Baseline 2 when weak labels are added back to help training. The boosts manifested in Baseline 4 and 5 indicate the advantage of pseudo labels and duration attention respectively.

\section{Variance in One Split}

\begin{figure}
\begin{center}
\includegraphics[width=0.8\linewidth]{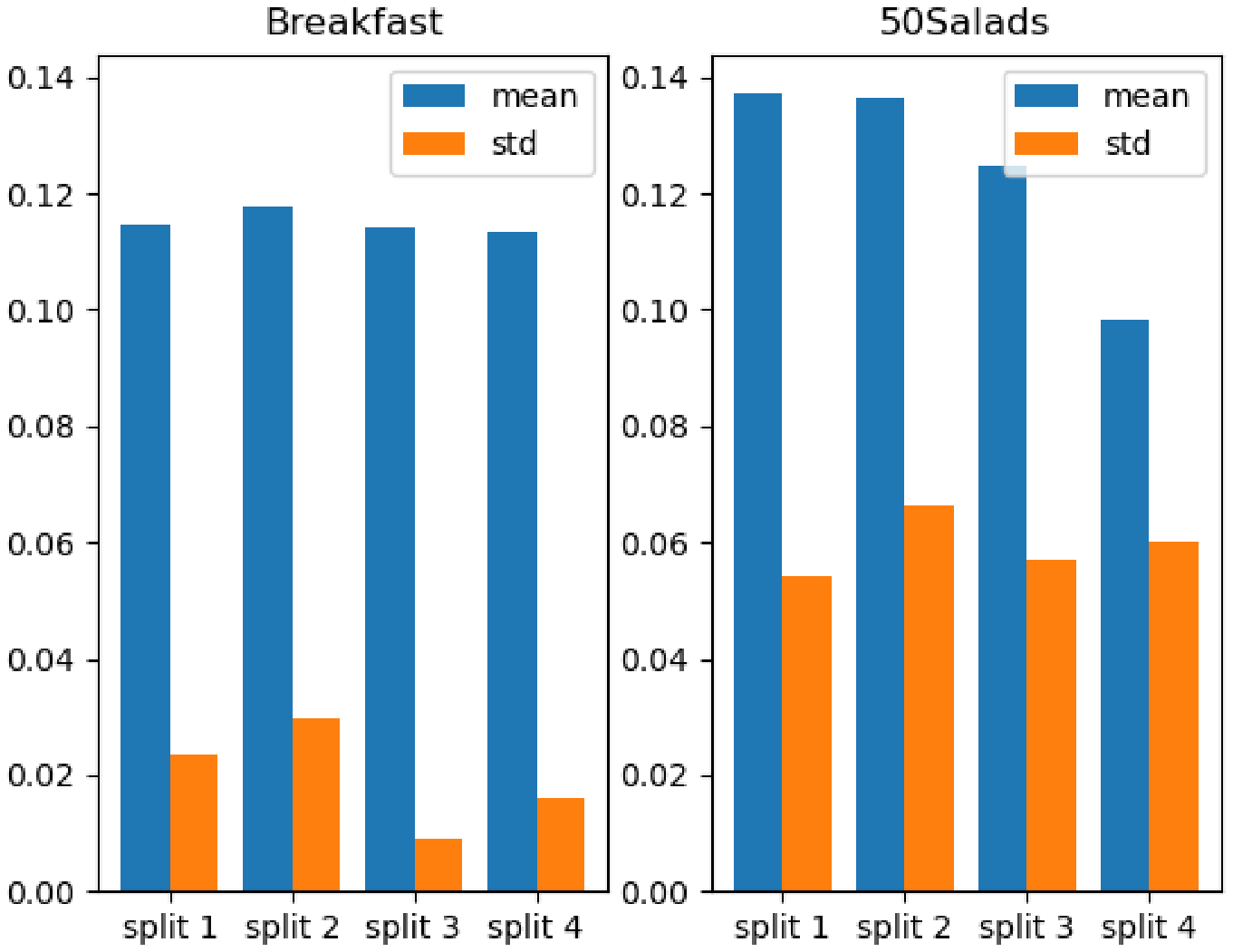}
\end{center}
\caption{Bar charts of means and standard deviations in each split.}
\label{fig:barchart}
\end{figure}

To further prove the randomness of our data choice and observe the variance in one split, we run 10 times on each split and plot the means and standard deviations. Here we use observation of 30\% and prediction of 20\% and use Sener's method as the backbone. As shown in Figure~\ref{fig:barchart}, we can see standard deviations of 50Salads are higher than those of Breakfast. The Reason may be that 50Salads has fewer videos, which is more unstable.

\section{Visualization of Attention Scheme}

\begin{figure}
\begin{center}
\includegraphics[width=0.8\linewidth]{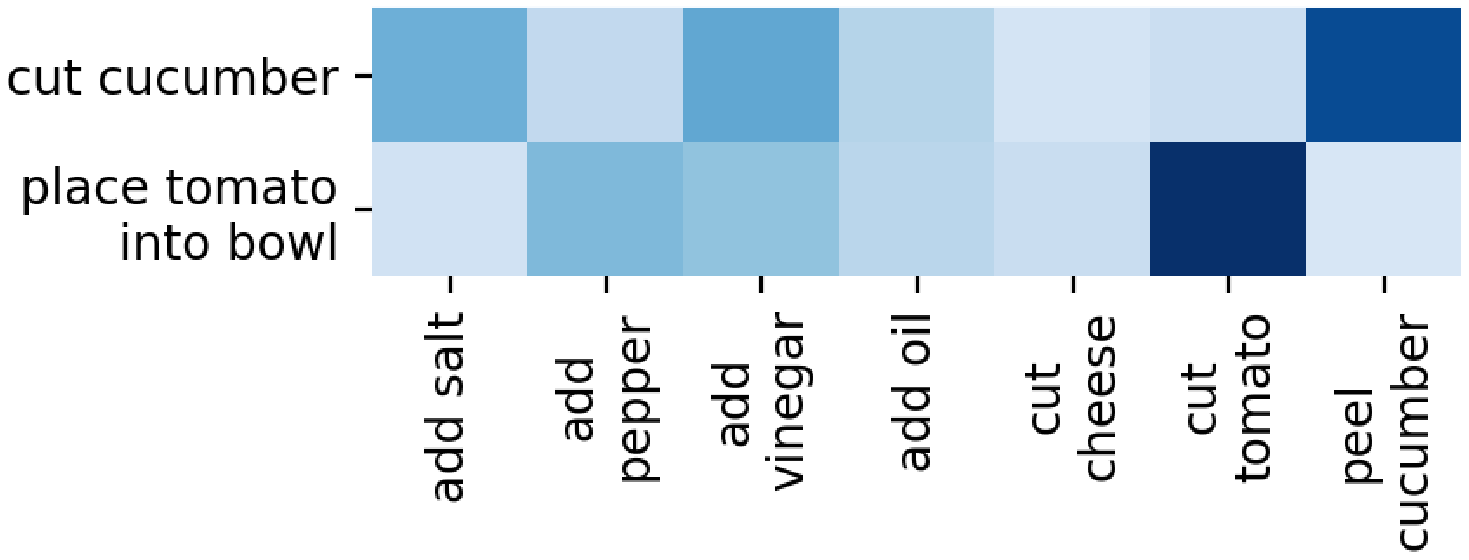}
\end{center}
\caption{Heat map of attention score between current predicted action and observed actions. x-axis is observed actions and y-axis is predicted actions. Deeper colour indicates higher attention score.}
\label{fig:heatmap}
\end{figure}

We use a heat map (Figure~\ref{fig:heatmap}) to further illustrate the advantage of our duration attention scheme. Take a video in 50Salads as an example, we track the attention score between current predicted action and observed actions. In the heat map, it's obvious that the correlation between ``cut cucumber'' and ``peel cucumber'' as well as the correlation between ``place tomato into bowl'' and ``cut tomato'' are the highest, which indicates that more relevant actions have more influence on current action duration.

\section{Full-Weak Split}

We vary the proportion of fully-labelled data in the training set in Table~\ref{tb:split} and observe that by increasing the proportion of fully-labelled data (the total amount of data is fixed), the performance gets progressively closer to the fully supervised model.
For the RNN and Ke's model, we are able to exceed the performance of the fully supervised model, though this is largely due to their poor baseline performance even with 100\% of the training data fully supervised.  It is likely that these models, being simpler and having fewer parameters, require a smaller proportion of fully-labelled data.
For a larger model like that of Sener, 25\% / 30\% of the data is not sufficient to match the fully-supervised performance. 

We omit experiments for RNN with split 25\% on Breakfast and split 30\% on 50Salads because MoC with smaller splits already exceeds fully-supervised results.  

\begin{table}[H]
\caption{MoC on different full vs. weak data splits. Percentages indicate the proportion of fully-labelled data in the training set. RNN and Sener use our full model in the weakly-supervised setting,\emph{i.e.} with duration attention while Ke, as a one-shot method, does not have duration attention.  $^*$100\% indicates the original fully-supervised model (also without duration attention).}
\begin{center}
\small
\begin{tabular}{|c|c|c|c|c|c|c|c|c|}
\hline
\quad & \multicolumn{4}{c|}{Breakfast} & \multicolumn{4}{c|}{50Salads} \\
\hline
\quad & 5\% & 15\% & 25\% & 100\%$^*$ & 10\% & 20\% & 30\% & 100\%$^*$ \\
\hline
RNN & 11.03 & 12.55 & \myslbox & 9.19 & 8.84 & 15.09 & \myslbox & 12.97 \\
\hline
Ke & 12.40 & 13.75 & 17.45 & 14.24 & 11.73 & 15.66 & 20.00 & 16.34 \\
\hline
Sener & 11.90 & 12.78 & 17.22 & 19.50 & 14.63 & 16.24 & 17.37 & 30.60 \\
\hline
\end{tabular}
\end{center}
\label{tb:split}
\end{table}

\section{Memory Complexity Analysis}

A simple comparison of memory complexity (expressed by the number of hyperparameters) of three baseline models (with fully-supervised setting) and our full model with adaptive refinement is shown in Table~\ref{tb:complexity}. Not surprisingly, the more complicated model has more hyperparameters. The number of hyperparameters of our full model is approximately two times the corresponding backbone's, which is in accordance with our intuition that the primary and conditional module is similar and are both based on the backbone. We omit time complexity analysis because it is not comparable between fully- and weakly-supervised models.

\begin{table}[H]
\caption{Memory complexity analysis.}
\begin{center}
\begin{tabular}{|c|c|c|c|c|c|}
\hline
\quad & RNN & Ke & Sener & Ours with RNN & Ours with Sener's\\
\hline
50Salads & 3579944 & 8937316 & 36997429 & 7159894 & 75374068 \\
\hline
Breakfast & 3624546 & 9845839 & 40270080 & 7249098 & 81590790 \\
\hline
\end{tabular}
\end{center}
\label{tb:complexity}
\end{table}

\section{Visualized Result}

\begin{figure}[H]
\begin{center}
\includegraphics[width=0.6\linewidth]{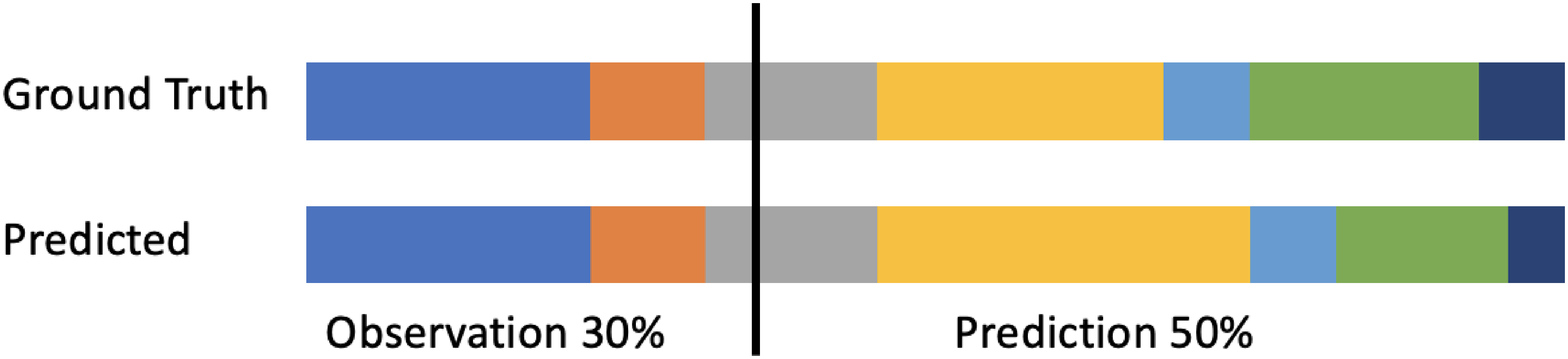}
\end{center}
   \caption{Visualized result for dense anticipation on Breakfast.}
\label{fig:visual}
\end{figure}

Figure~\ref{fig:visual} shows an example of anticipating 50\% of the sequence after observing 30\%, where each colour indicates an action. 
We can see that the action sequence is correct, but there are some errors in the predicted duration.

\section{Pseudo Codes}

Below are pseudo codes for linear refinement and adaptive refinement respectively.

\begin{algorithm}
\caption{Linear Refinement}
\begin{algorithmic} 
\Require \text{initial model }$Prim, Cond$; $\mathcal{W}, \mathcal{F}$; \text{Epoch} $N_1, N_2$; $\alpha$, \text{decay parameter} $d$
\State $\text{Step 1:}$
\For {$n = 1\text{ to }N_1$} 
\State $pseudo\_label \leftarrow Cond(\mathcal{F})$
\State $L \leftarrow Loss(ground\_truth, pseudo\_label)$
\State \text{Update} $Cond$ \text{by minimizing} $L$
\EndFor
\State \text{Fix} $Cond$
\State $\text{Step 2:}$
\For {$n = 1\text{ to }N_2$} 
\State $predicted\_label \leftarrow Prim(\mathcal{F})$
\State $L_1 \leftarrow Loss(ground\_truth, predicted\_label)$
\State $predicted\_label \leftarrow Prim(\mathcal{W})$
\State $pseudo\_label \leftarrow Cond(\mathcal{W})$
\State $refined\_label \leftarrow predicted\_label^{\frac{1}{\alpha+1}}*pseudo\_label^{\frac{\alpha}{\alpha+1}}$
\State $L_2 \leftarrow Loss(refined\_label, predicted\_label)$
\State \text{Update} $Prim$ \text{by minimizing} $L_1+L_2$
\State $\alpha \leftarrow d*\alpha$
\EndFor
\end{algorithmic}
\end{algorithm}

\begin{algorithm}[t!]
\caption{Adaptive Refinement}
\begin{algorithmic} 
\Require \text{initial model }$Prim, Cond, Refine$; $\mathcal{W}, \mathcal{F}$; \text{Epoch} $N_1, N_2, N_3$
\State $\text{Step 1:}$
\For {$n = 1\text{ to }N_1$} 
\State $pseudo\_label \leftarrow Cond(\mathcal{F})$
\State $L \leftarrow Loss(ground\_truth, pseudo\_label)$
\State \text{Update} $Cond$ \text{by minimizing} $L$
\EndFor
\State \text{Fix} $Cond$
\State $\text{Step 2:}$
\For {$n = 1\text{ to }N_2$} 
\State $predicted\_label \leftarrow Prim(\mathcal{F})$
\State $pseudo\_label \leftarrow Cond(\mathcal{F})$
\State $refined\_label \leftarrow Refine(predicted\_label, pseudo\_label)$
\State $L_1 \leftarrow Loss(ground\_truth, predicted\_label)$
\State $L_2 \leftarrow Loss(refined\_label, predicted\_label)$
\State \text{Update} $Prim, Refine$ \text{by minimizing} $L_1+L_2$
\EndFor
\State $\text{Step 3:}$
\For {$n = 1\text{ to }N_3$} 
\State $predicted\_label \leftarrow Prim(\mathcal{F})$
\State $pseudo\_label \leftarrow Cond(\mathcal{F})$
\State $refined\_label \leftarrow Refine(predicted\_label, pseudo\_label)$
\State $L_1 \leftarrow Loss(ground\_truth, predicted\_label)$
\State $L_2 \leftarrow Loss(refined\_label, predicted\_label)$
\State $predicted\_label \leftarrow Prim(\mathcal{W})$
\State $pseudo\_label \leftarrow Cond(\mathcal{W})$
\State $refined\_label \leftarrow Refine(predicted\_label, pseudo\_label)$
\State $L_3 \leftarrow Loss(refined\_label, predicted\_label)$
\State \text{Update} $Prim, Refine$ \text{by minimizing} $L_1+L_2+L_3$
\EndFor
\end{algorithmic}
\end{algorithm}
